\documentclass[journal]{IEEEtran}
\IEEEoverridecommandlockouts 

\usepackage{cite}
\usepackage{amsmath,amssymb,amsfonts}
\usepackage{algorithmic}
\usepackage{graphicx}
\usepackage{textcomp}
\usepackage{xcolor}
\usepackage{booktabs}
\def\BibTeX{{\rm B\kern-.05em{\sc i\kern-.025em b}\kern-.08em
    T\kern-.1667em\lower.7ex\hbox{E}\kern-.125emX}}

\usepackage{epsfig}
\usepackage{bm}
\usepackage{enumerate}
\usepackage{multirow}
\usepackage{subcaption}
\usepackage{array}
\newcolumntype{H}{>{\setbox0=\hbox\bgroup}c<{\egroup}@{}} 

\usepackage{hyperref}
\hypersetup{pagebackref=true,breaklinks=true,colorlinks,bookmarks=false,urlcolor=blue}

\usepackage{afterpage}
\usepackage{tikz}

\newcommand{\stufig}[5]                                    
{
	\begin{figure}[#5]
		\begin{center}
			\includegraphics[#1]{#2}
			\caption{#3}
			\label{#4}
		\end{center}
		\vspace{-.5cm}
	\end{figure}
}

\newcommand{\stufigstar}[5]                                
{
	\begin{figure*}[#5]
		\begin{center}
			\includegraphics[#1]{#2}
			\caption{#3}
			\label{#4}
		\end{center}
		\vspace{-.5cm}
	\end{figure*}
}

\newenvironment{stusubfig*}[1]
{
	\begin{figure*}[#1]
		\begin{center}
		}
		{
		\end{center}
	\end{figure*}
}

\begin{document}

\title{Real-Time Highly Accurate Dense Depth on a Power Budget using an FPGA-CPU Hybrid SoC}

\author{
	\begin{tabular}{c@{\hskip 0.85cm}c@{\hskip 0.85cm}c@{\hskip 0.85cm}c@{\hskip 0.85cm}c}
    Oscar Rahnama & Tommaso Cavallari & Stuart Golodetz & Alessio Tonioni & Thomas Joy \\
    & Luigi Di Stefano & Simon Walker & Philip H.\ S.\ Torr
  \end{tabular}%
  \thanks{Correspondence: {\texttt{\{oscar@robots.ox.ac.uk\}}}}
  \thanks{O.\ Rahnama is with the University of Oxford and FiveAI Ltd.}
  \thanks{T.\ Joy and P.\ Torr are with the University of Oxford.}
  \thanks{A.\ Tonioni and L.\ Di Stefano are with the University of Bologna.}
  \thanks{T.\ Cavallari, S.\ Golodetz and S.\ Walker are with FiveAI Ltd.}
  \thanks{Work done whilst A.\ Tonioni was visiting the University of Oxford.}
}

\maketitle

\newcommand\copyrighttext{%
    \footnotesize \textcopyright{} 2019 IEEE.
    Personal use of this material is permitted.
    Permission from IEEE must be obtained for all other uses, in any current or future media, including reprinting/republishing this material for advertising or promotional purposes, creating new collective works, for resale or redistribution to servers or lists, or reuse of any copyrighted component of this work in other works.
    The published version of this paper has DOI: \href{https://dx.doi.org/10.1109/TCSII.2019.2909169}{10.1109/TCSII.2019.2909169}}
  \newcommand\copyrightnotice{%
    \begin{tikzpicture}[remember picture,overlay]
    \node[anchor=south,yshift=5pt] at (current page.south) {\fbox{\parbox{\dimexpr\textwidth-\fboxsep-\fboxrule\relax}{\copyrighttext}}};
    \end{tikzpicture}%
}

\copyrightnotice

\begin{abstract}
Obtaining highly accurate depth from stereo images in real time has many applications across computer vision and robotics, but in some contexts, upper bounds on power consumption constrain the feasible hardware to embedded platforms such as FPGAs.
Whilst various stereo algorithms have been deployed on these platforms, usually cut down to better match the embedded architecture, certain key parts of the more advanced algorithms, e.g.\ those that rely on unpredictable access to memory or are highly iterative in nature, are difficult to deploy efficiently on FPGAs, and thus the depth quality that can be achieved is limited.
In this paper, we leverage a FPGA-CPU chip to propose a novel, sophisticated, stereo approach that combines the best features of SGM and ELAS-based methods to compute highly accurate dense depth in real time.
Our approach achieves an 8.7\% error rate on the challenging KITTI 2015 dataset at over 50 FPS, with a power consumption of only 5W.
\end{abstract}


\IEEEpeerreviewmaketitle

Obtaining information about the 3D structure of a scene is important for many computer vision and robotics applications, e.g.\ 3D scene reconstruction \cite{whelan2016, prisacariu2017, golodetz2018}, camera relocalisation \cite{shotton2013, cavallari2017, cavallari2018}, navigation and obstacle avoidance \cite{hicks2013}.
Often, this information will be obtained in the form of a depth image, and various options for acquiring such images exist.
Passive approaches, which rely only on one or more image sensors, are popular due their low cost, low weight and size, lack of active/moving components, ability to work at longer ranges, deployability in a wider range of operating environments and lack of interference.
Among them, binocular stereo relies on a pair of synchronised cameras to acquire the same scene from two different points of view.
Given the two frames, a dense and reliable depth map can be computed by finding correspondences between the pixels in the two images \cite{scharstein2002taxonomy}.
State-of-the-art algorithms for this problem usually rely on costly global image optimisations or on massive convolutional neural networks that involve significant computational costs, making them hard to deploy on resource-limited systems such as embedded devices \cite{tippetts2016review}.

Two popular solutions offering a good trade-off between speed and accuracy are Semi-Global Matching (SGM) \cite{hirschmuller2008stereo} and ELAS \cite{geiger2010efficient}.
SGM computes initial matching hypotheses by comparing patches around pixels in the left and right images, then approximates a costly image-wide smoothness constraint with the sum of several directional minimizations over the disparity range.
By contrast, ELAS first identifies a set of sparse but reliable correspondences to provide a coarse approximation of the scene geometry, then uses them to define slanted plane priors that guide the final dense matching stage.
We propose a novel stereo pipeline that efficiently combines the predictions of these two algorithms, achieving high accuracy and overcoming some of the limitations of each algorithm.
First, we use multiple passes of a fast SGM variant~\cite{rahnama2018fpt}, left-right consistency checking and decimation to obtain a sparse but reliable set of correspondences.
Then, we use these as the support points for ELAS to obtain disparity priors from slanted planes.
Finally, we incorporate these disparity priors into a final SGM-based optimization (again based on~\cite{rahnama2018fpt}) to achieve dense predictions with high accuracy.

Our pipeline targets not only accuracy, but also speed, aiming for real-time execution (30 fps) on an embedded platform.
Recent works have deployed SGM successfully in real time both on multi-core CPUs~\cite{gehrig2010real} and GPUs~\cite{banz2011real,hernandez2016embedded}, but in real-world scenarios, power constraints often force us to rely on low-power devices like FPGAs.
The development of reliable stereo pipelines for FPGAs is an active research field \cite{perri2018design, rahnama2017real, zhang2011real, perri2018stereo, ttofis2014high, dehnavi2017fpga, zha2016real, tippetts2016review}, with recent works proposing FPGA-friendly variants of SGM \cite{gehrig2009real, banz2010real, mattoccia2015passive, honegger2014real, wang2015real, hernandez2016embedded} or ELAS \cite{rahnama2018real}.
However, FPGA implementations of stereo algorithms usually perform some kind of approximation to deal with the limited resources available and to traverse the pixels in raster order.

We show how some of the intrinsic limitations of a pure FPGA-based implementation can be mitigated by appropriately leveraging a new-generation hybrid system on a chip (SoC), e.g.\ the Xilinx ZCU104, which combines both an ARM processor and an FPGA, with shared direct memory access, into a single chip.
Recently, several works have explored the deployment of stereo methods on such platforms: both \cite{honegger2014real} and \cite{perri2018stereo} use the CPU mainly for handling communication and controlling peripherals, while \cite{rahnama2018real} actively leverages the CPU to execute iterative steps that would be infeasible on an FPGA (e.g.\ Delaunay triangulation).
Similar to \cite{rahnama2018real}, we propose to actively use the elaboration capability of the built-in CPU to handle I/O and to execute part of the ELAS pipeline, while deploying all the other elaboration blocks on the FPGA.
We show how our pipeline outperforms previously published works by achieving an 8.7\% error rate on the challenging KITTI 2015 dataset \cite{Menze2015ISA,Menze2018JPRS}, while still operating with real-time performance and low power consumption.

	\begin{figure*}[!t]
		\centering
		\includegraphics[width=\linewidth]{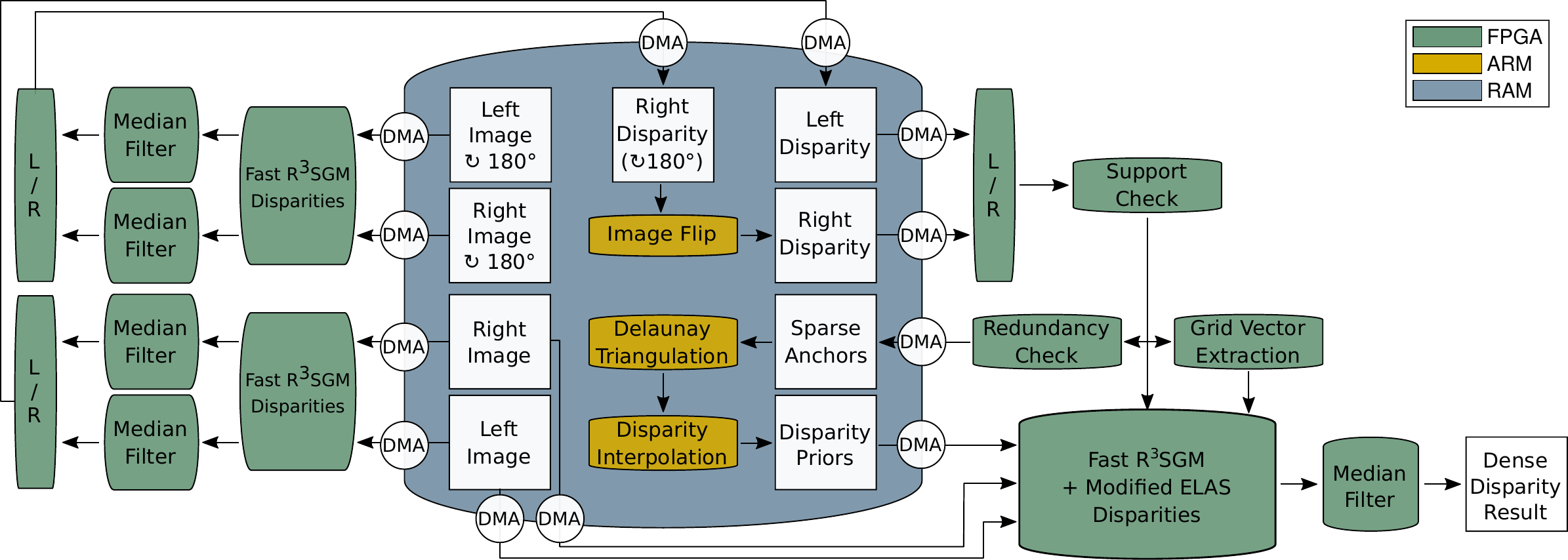}
		\caption
		[
		System-level overview of heterogenous approach to depth computation
		]
		{
			\textbf{Overview of our approach}.
			First, we use Fast R$^3$SGM (see \S\ref{subsec:fastr3}) to compute disparity images for the input stereo pair (in raster and reverse-raster order).
			We then flip the right result and perform a left-right consistency check to obtain an accurate but sparse disparity map for the left input image (see \S\ref{subsec:lrcheck}).
			Next, as ELAS \cite{geiger2010efficient} does, we perform \emph{support checking} (see \S\ref{subsec:supportcheck}) to remove points whose disparities appear abnormal relative to neighbouring pixels: this yields a sparser support point image that contains only points with confident disparities.
			This support point image is subsequently used in multiple ways:
			(i) it is further sparsified via a \emph{redundancy check}, producing sparse anchors that are then
			used to generate plane disparity priors through a \emph{triangulation} and \emph{interpolation} process (see \S\ref{subsec:redundancycheck}); 
			(ii) it is split into a grid where, for each grid cell, a binary vector representing the set of viable disparities is computed (see \S\ref{subsec:elas}).
			Finally, the support point image is combined with the outputs of (i) and (ii) in a \emph{disparity optimization} that combines R$^3$SGM and ELAS to produce a dense disparity image (see \S\ref{subsec:combinedopt}).
			We then median filter this image for robustness to produce the final result.
			\vspace{-\baselineskip}
		}
		\label{fig:pipeline}
	\end{figure*}
   
\section{Method}

Our overall pipeline is shown in Figure~\ref{fig:pipeline}.
It consists of several different components which we describe in the subsections that follow.
The system leverages both parts of the FPGA-CPU hybrid SoC to achieve optimal results.
Tasks that are very data-intensive, but which access that data in a predictable manner, are run on dedicated FPGA accelerators to benefit from their parallel processing capability.
In addition, they can take advantage of the FPGA accelerators' internal ability to pipeline data so that multiple inputs are processed together in staggered fashion.
Tasks that are very dynamic and unpredictable, which often involve many unforeseen or random accesses to external memory, are run on the CPU, since they benefit both from the significantly faster clock frequency of the CPU and its ability to access memory in constant time (CPU memory accesses can be sped up via appropriate use of the cache).
To minimize the amount of FPGA resources used by our method, as well as allow the deployment of the design on a real platform, we reuse some accelerators whilst buffering intermediate results in RAM.
We will detail which blocks are reused in our final design in the rest of this section.

\subsection{Sparse Disparity Computation}

\subsubsection{Fast R$^3$SGM}
\label{subsec:fastr3}

Initially, we use a modified version of R$^3$SGM~\cite{rahnama2018fpt} (a memory-efficient adaptation of classic SGM~\cite{hirschmuller2008stereo} to FPGAs), which we call Fast R$^3$SGM, to compute disparity images for input stereo pairs.
The original version of R$^3$SGM aggregated contributions to the disparity of each pixel along four different scanlines: three above the pixel, and one to the left.
However, as mentioned in~\cite{rahnama2018fpt},  using the left scanline severely limits the overall throughput of the system (one disparity value is output every three clock cycles) due to a blocking dependency between immediately successive pixels.
To avoid this, we modify the approach to use only the scanlines above the pixel, allowing us to output one disparity per clock cycle.
The mild loss in accuracy this causes is more than compensated for by the improvements yielded by the rest of our pipeline.

\begin{figure}[th]
	\centering
	\begin{subfigure}{.30\linewidth}
		\centering
		\includegraphics[height=\linewidth]{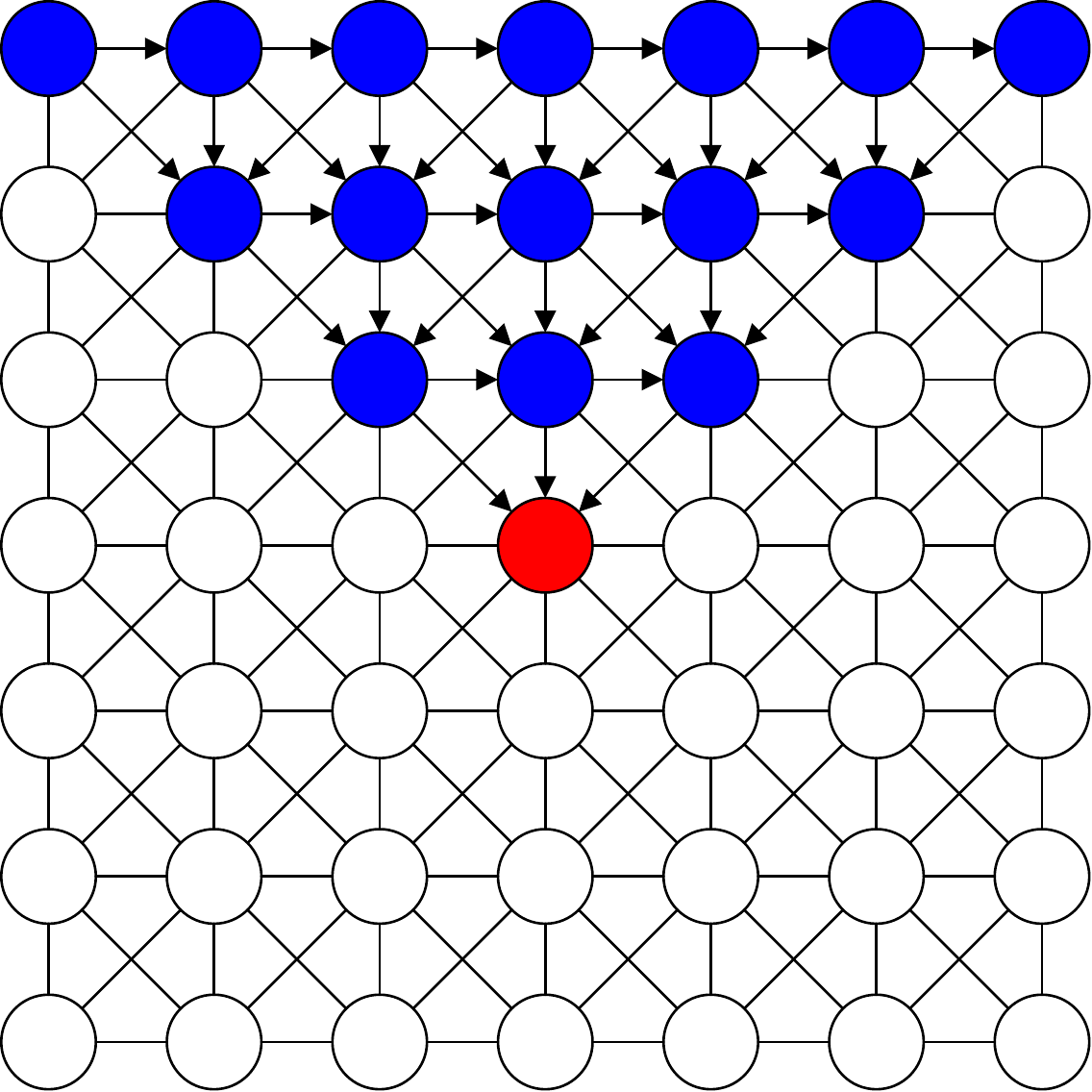}
		\caption{}
		\label{fig:r2fwd}
	\end{subfigure}%
	\hspace{10mm}%
	\begin{subfigure}{.30\linewidth}
		\centering
		\includegraphics[height=\linewidth]{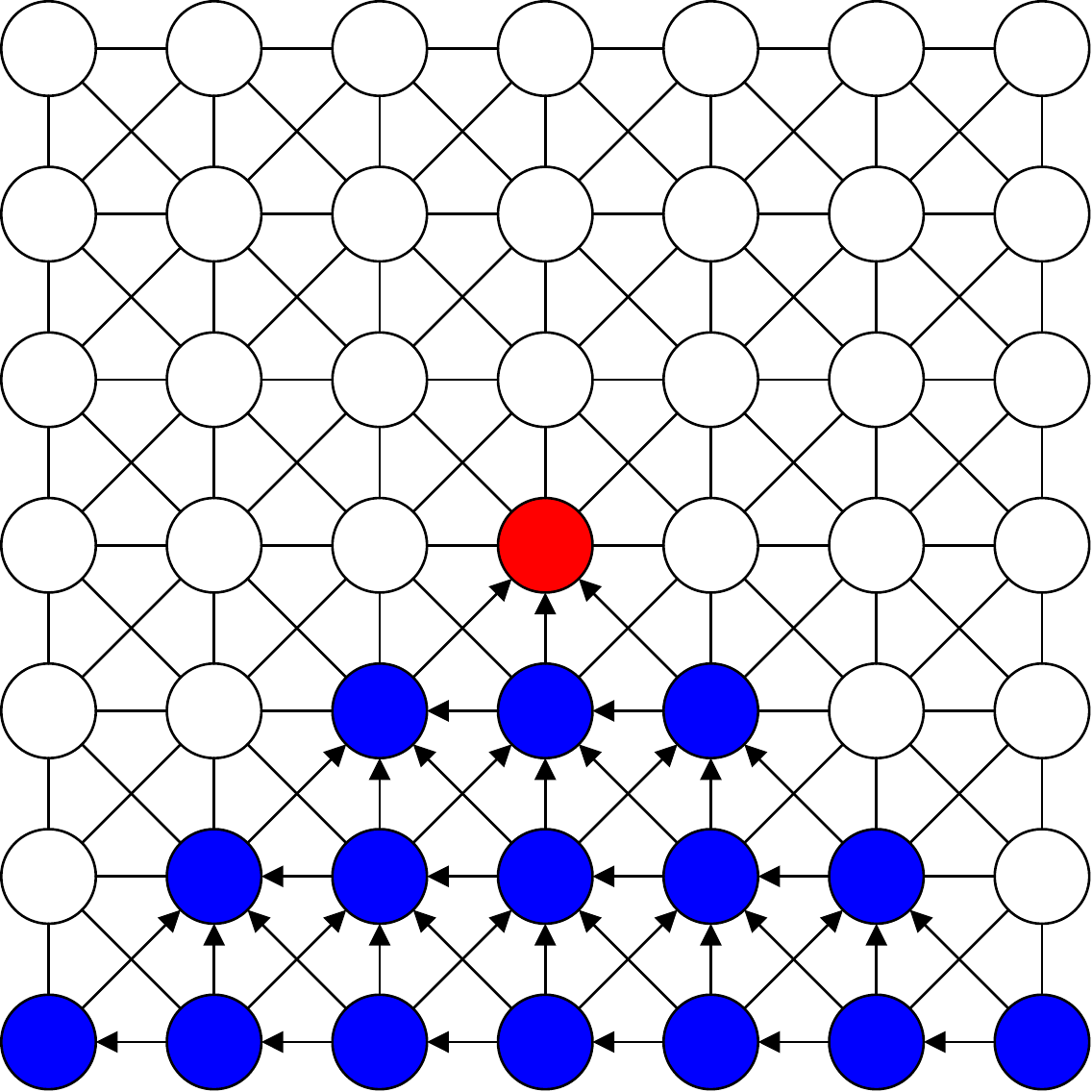}
		\caption{}
		\label{fig:r2bck}
	\end{subfigure}%
	\caption
	[
	Comparing bias in raster and reverse raster passes of Fast R$^3$SGM
	]
	{
		Comparing the implicit biases that exist in raster and reverse-raster passes of R$^3$SGM.
    The region of influence used in the original version of Fast R$^3$SGM contains only pixels above the pixel of interest (\subref{fig:r2fwd}), and as such the disparity value computed for that pixel is unaffected by the pixels below it. 
		The opposite holds true when performing Fast R$^3$SGM in reverse-raster order (\subref{fig:r2bck}). 
	}
	\label{fig:r2bias}
\end{figure}

We process each input pair twice: once in raster order, and once in reverse-raster order, yielding two disparity images overall.
The advantage of this is that, as illustrated in Figure~\ref{fig:r2bias}, the raster and reverse-raster passes of R$^3$SGM will base the disparity for each pixel on the disparities of pixels in different regions of influence.
By comparing the results output by both these separate passes, we can identify pixels for which the difference in bias caused a significant change in the  resulting output value.
These can then be removed so that only pixels that maintain the same value, regardless of bias, are retained. 
Through this type of consistency check, the confidence and accuracy of the results can be improved.

\begin{figure}
	\centering
	\includegraphics[width=\linewidth]{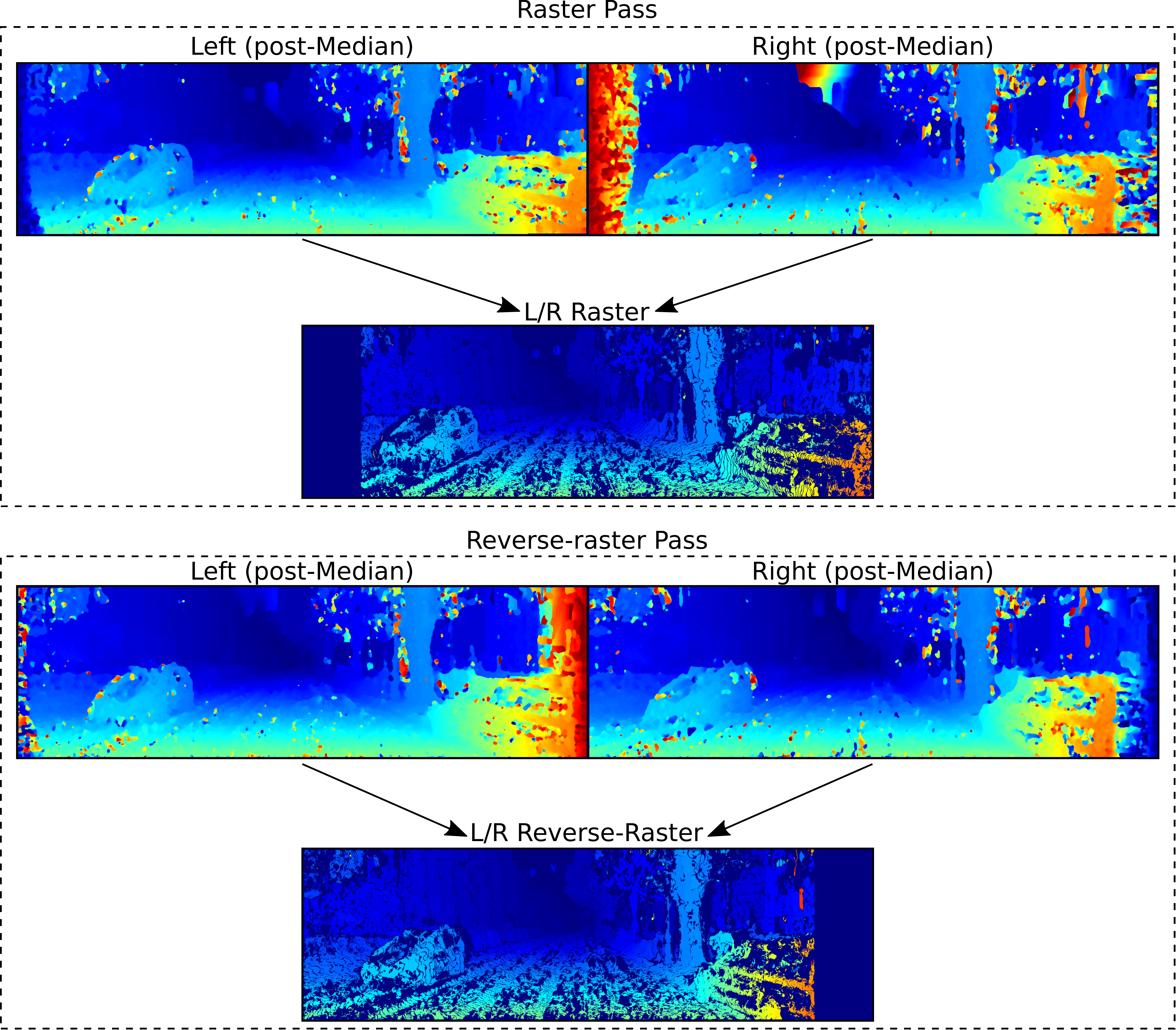}
	\caption
	[
	Raster and Reverse Raster Passes
	]
	{
    An example showing the effects of performing L/R consistency checks after the initial raster and reverse-raster passes of Fast R$^3$SGM and the subsequent median filtering. Note that in the reverse-raster case, we have flipped the images for easier comparison with the raster case.
	}
	\label{fig:rast_rev_images}
\end{figure}

In our implementation, we deploy a single instance of the Fast R$^3$SGM block, together with the associated median filtering and L/R consistency checking blocks.
We first feed the blocks with the raster-order stereo pair, then with the reverse-raster-order pair, storing the disparities resulting from each pair back into RAM between the computations.
Example images resulting from this two-pass process are depicted in Figure~\ref{fig:rast_rev_images}.
For further architectural details of the internal structure of the main blocks, we refer the interested reader to~\cite{rahnama2018fpt}.

\subsubsection{Consolidating Consistency Checking}
\label{subsec:lrcheck}

Each pass of Fast R$^3$SGM outputs a disparity map that has been checked for consistency using the first input as the reference image \cite{rahnama2018fpt}.
The raster pass outputs a disparity map for the left input image; the reverse-raster pass outputs one for the (reversed) right input image.
Due to the streaming nature of the disparity computation, however, the results suffer from a raster or reverse-raster scan bias, i.e.\ the disparity value of any given pixel is encouraged to be similar to those computed before it.
To reconcile the inconsistencies between these two disparity maps, we perform a further left-right consistency check, which yields an accurate but sparse disparity map for the left input image as its result (see Figure~\ref{fig:pipeline}).
The memory access pattern of such a process is problematic, however, as the first pixels in the left disparity map need to be checked against the last pixels in the right disparity map.
To overcome this problem, we first reverse the latter image on the CPU (since this is an inherently sequential process, it benefits from the higher clock rate provided by the ARM core), then perform a standard left-right consistency check (on the programmable logic).

\subsection{Generation of Priors}

Using the sparse disparity map output by the consolidating consistency check, we adapt the ELAS method described in \cite{geiger2010efficient} to generate priors that can be fed into a combined disparity optimization process (see \S\ref{subsec:combinedopt}) to produce a more accurate and dense final result.
The prior generation process begins by taking the disparity map produced by \S\ref{subsec:lrcheck} as input and producing a support point image (see \S\ref{subsec:supportcheck}) containing sparse but confident disparities.
The support points are then fed to two more blocks before being used by the final disparity optimization process: (i) a redundancy checking and disparity prior generation block, which first computes a sparse anchor points image and then triangulates such anchors to generate disparity priors for all pixels in the image (see \S\ref{subsec:redundancycheck}); and (ii) a grid vector extraction block that divides the support points image into a grid and then determines the set of possible disparities for each cell (see \S\ref{subsec:elas}).

\subsubsection{Support Checking}
\label{subsec:supportcheck}

\begin{figure}
	\centering
	\includegraphics[width=\linewidth]{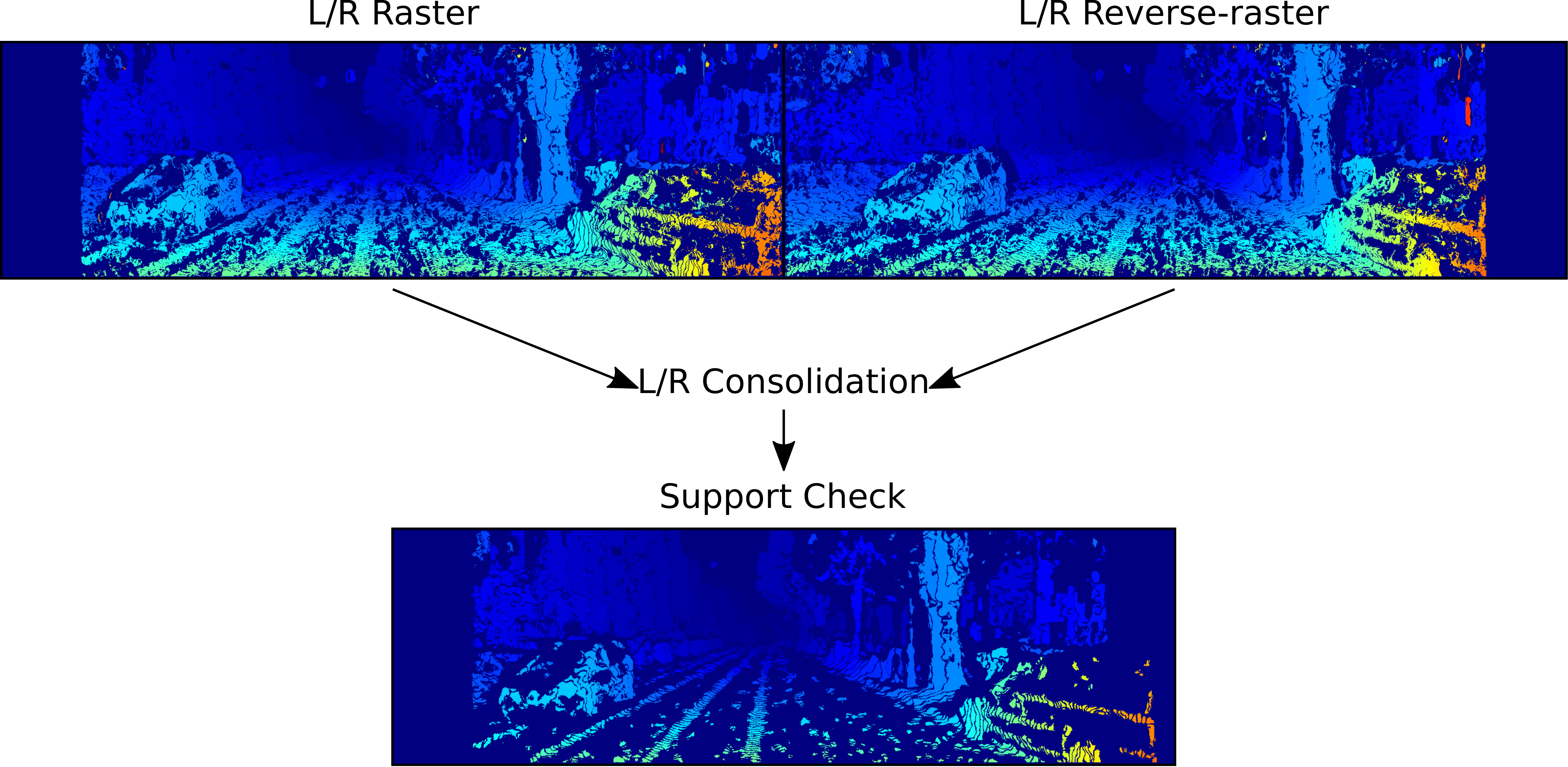}
	\caption
	[
	Example image of support points
	]
  {
	The results of performing a consolidating consistency check and subsequent support check on the raster and reverse-raster outputs from Figure~\ref{fig:rast_rev_images}.
	}
	\label{fig:support_images}
\end{figure}

To produce the support point image (Figure~\ref{fig:support_images}), we filter the sparse disparity map to remove any pixels whose disparities are not sufficiently supported by the pixels in their immediate neighbourhoods (in practice, a square window centred on each pixel).
For a pixel to be considered ``supported'', there must exist, in its neighbourhood, another predefined number of pixels that have very similar disparity values (e.g.\ at least $10$ pixels within a $5 \times 5$ window that differ by less than $5$ from that of the centre pixel).
The disparities of all other pixels are marked as invalid.
The resulting support point image will evidently be sparser than the original disparity map, since we have kept only those pixels about whose disparities we can be reasonably confident.

\subsubsection{Redundancy Checking and Disparity Prior Generation}
\label{subsec:redundancycheck}

\begin{figure}[!t]
	\centering
	\includegraphics[width=\linewidth]{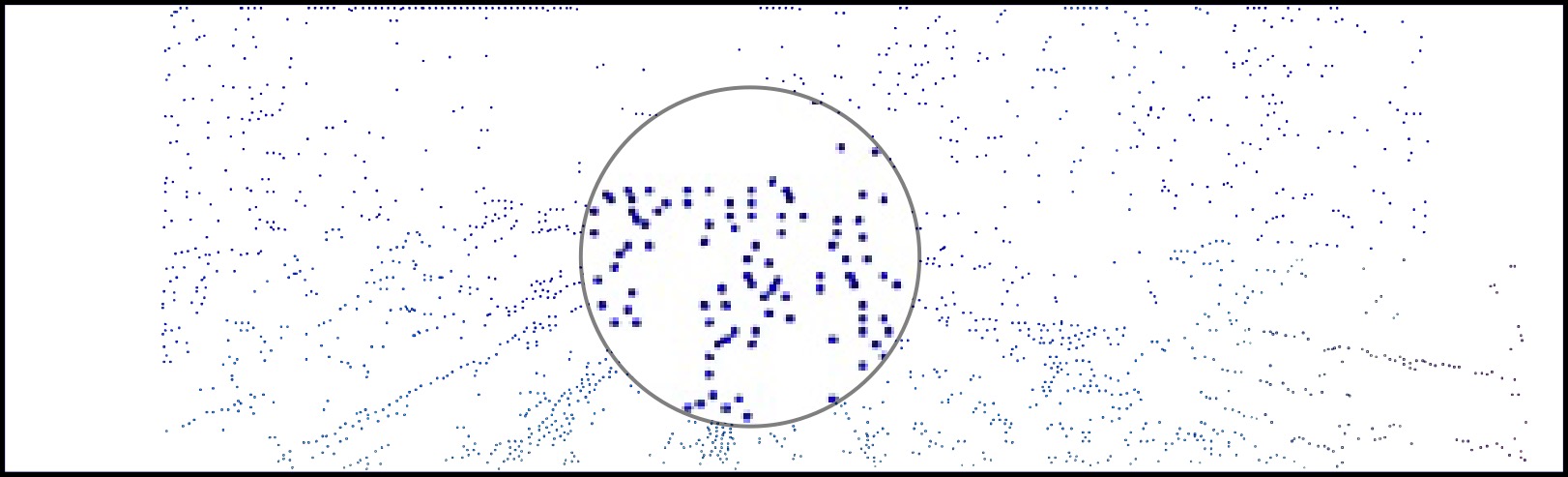}
	\caption
	[
	Example of sparse anchor points image
	]
	{
    The results of performing a redundancy check on the support point image from Figure~\ref{fig:support_images}.
    Note the sparseness of the anchor points, which significantly speeds up the subsequent Delaunay triangulation process.
    A portion of the image has been magnified for better visibility.
	}
	\label{fig:sparseanchors}
\end{figure}

To produce the anchor image (see Figure~\ref{fig:sparseanchors}), we further sparsify the support point image from \S\ref{subsec:supportcheck} by processing it in raster order and invalidating any pixel whose disparity has already been seen within a window behind and above it.
Unlike \cite{rahnama2018real}, which for each pixel $(x,y)$ used a window of
\begin{center}
	\footnotesize
	\begin{tabular}{c}
		$\{(x,y-\delta_y) : 0 < \delta_y \le 2K\} \cup \{(x-\delta_x,y) : 0 < \delta_x \le 2K\},$
	\end{tabular}
	\normalsize
\end{center}
where $K$ was set to $5$, which only encompassed points in the same row or same column as the pixel being processed, here we use a larger window of

\begin{center}
	\footnotesize
	\begin{tabular}{cl}
		& $\{(x+\delta_x,y+\delta_y) : -K \le \delta_x \le K, -2K \le \delta_y < 0\}$ \\
		$\cup$ & $\{(x-\delta_x,y) : 0 < \delta_x \le K\}$.
	\end{tabular}
	\normalsize
\end{center}

\noindent This has the effect of creating a sparser anchor image than that used in \cite{rahnama2018real}, significantly speeding up the subsequent Delaunay triangulation process.
Whilst this inevitably reduces the granularity of the generated triangles, its impact on the quality of the subsequent depth priors is minor, as shown in \cite{rahnama2018real}.
As the Delaunay triangulation process was additionally shown to be a key bottleneck, the advantage of reducing the number of points to triangulate (thus reducing CPU processing time) outweighs the marginal benefit in accuracy obtained with more fine-grained triangles.

\begin{figure}[!b]
	\centering
	\includegraphics[width=\linewidth]{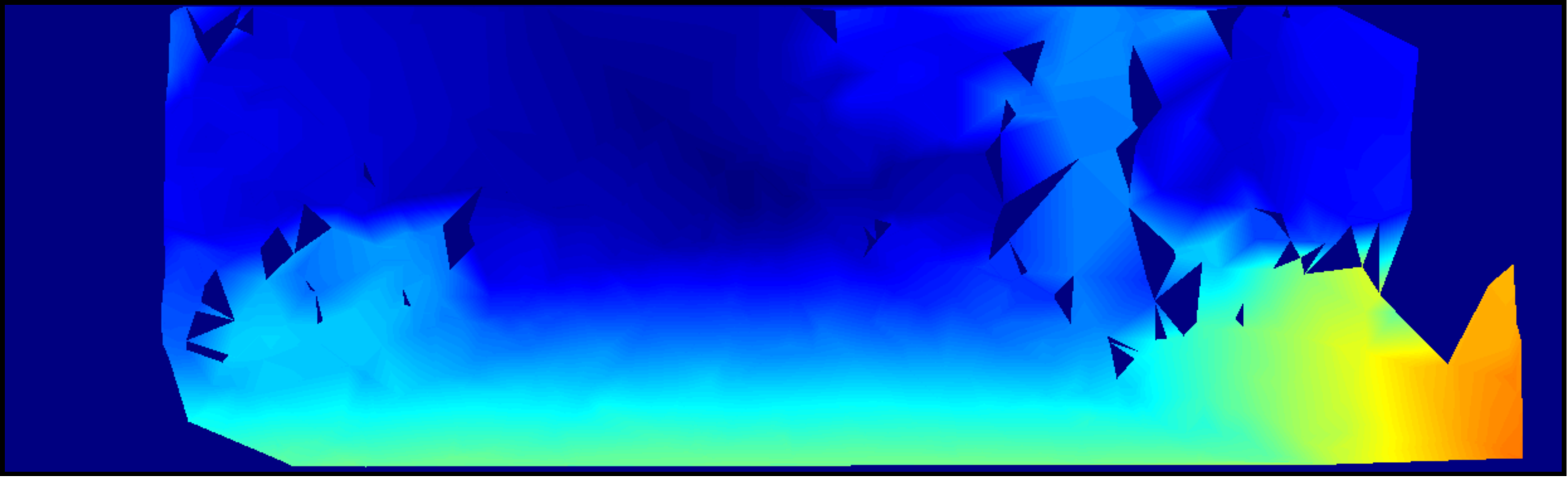}
	\caption
	[
	Example of plane prior image
	]
	{
   The plane priors produced by constructing a Delaunay triangulation based on the sparse anchor points in Figure~\ref{fig:sparseanchors}, and then linearly interpolating the disparities within each triangle.
	}
	\label{fig:planepriors}
\end{figure}

Finally, to produce the disparity priors, we first move the anchor points image back to RAM, then perform a Delaunay triangulation of those points, and finally compute the disparity of each non-anchor point located within one of the Delaunay triangles by interpolating the disparities of the triangle's vertices (depicted in Figure~\ref{fig:planepriors}).
The entirety of this process is performed by the CPU, since the triangulation and interpolation algorithms are inherently non-sequential in their memory access patterns, and can benefit from both the availability of memory caches and the higher speed of the ARM core.

\subsubsection{Grid Vector Extraction}
\label{subsec:elas}

The final input to the combined disparity optimization we describe in \S\ref{subsec:combinedopt} is a set of binary \emph{grid vectors} used to determine which disparities are suitable for each part of the image.
To produce such vectors, we first divide the support point image into a regular grid (with cells of size $50 \times 50$ in our implementation).
Then, for each cell, we find the valid disparity values within it, and store both those and their neighbouring disparities ($\pm 1$) into a binary grid vector for that cell.
See \cite{rahnama2018real} for more details.

\subsection{Combined Disparity Optimization}
\label{subsec:combinedopt}

Finally, we perform a combined disparity optimization that takes into account not only the original pair of input images, but also the plane priors, grid vectors and support points.
Essentially, we perform Fast R$^3$SGM, as in \S\ref{subsec:fastr3} (once again reusing the corresponding FPGA block), but first modifying the cost vectors of the pixels to take the various different priors we have available into account.

\begin{figure}[!t]
	\centering
	\includegraphics[width=\linewidth]{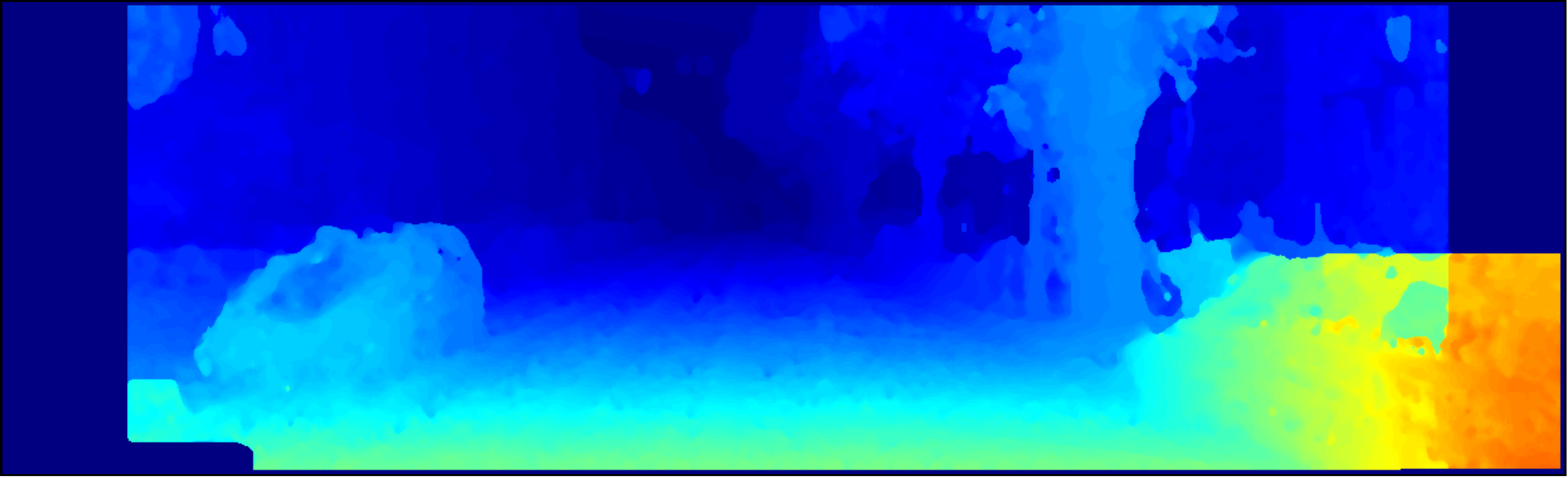}
	\caption
	[
	Example of final dense image
	]
	{
    The final dense disparity image for our running example, produced using the combined disparity optimisation described in Section~\ref{subsec:combinedopt}.
	} 
	\label{fig:finalimage}
\end{figure}

The disparities of the support points are fixed throughout and not recomputed.
Every cost vector element for a support point (each of which corresponds to a specific disparity) is set to a large arbitrary value, except for the element corresponding to the disparity of the support point, which is set to zero instead.
Through the Fast R$^3$SGM smoothing process, pixels near the support point will then naturally be encouraged to adopt disparities similar to that of the support point itself, with the influence of this effect attenuating with distance.
To take the disparity prior for each pixel into account, we decrease those elements of its cost vector that correspond to disparities close to the prior (more specifically, we superimpose a negative Gaussian over the cost vector, centred on the prior, and decrease the relevant elements within a certain radius accordingly).
To make use of the grid vectors, we set all elements of the cost vectors for the pixels within each grid cell that do not appear in the grid vector for that cell to an arbitrarily large value, thus strongly encouraging them not to be selected.
As with the effects of the support points, these cost vector modifications are similarly propagated by the Fast R$^3$SGM smoothing process.
At the end of this process, we perform a final median filter on the Fast R$^3$SGM result to further mitigate impulsive noise, ultimately yielding a dense, accurate disparity map as demonstrated in Figure~\ref{fig:finalimage}.

	\begin{table*}[!t]
		\centering
		\begin{tabular}{cccccccc}
			\toprule
			\textbf{Method} & \textbf{Background} & \textbf{Foreground} & \textbf{All} & \textbf{Density} & \textbf{Runtime (s)} & \textbf{Environment} & \textbf{Power Used (W)} \\
			\midrule
			Ours  & 7.2\% & 17.3\% & 8.7\% & 99.7\% & 0.019 & FPGA (Xilinx ZCU104) & $\approx$ 5 \\
			\midrule
			R$^3$SGM \cite{rahnama2018fpt} &  -- & -- & 9.9\% & 85.0\% & 0.014 & FPGA (Xilinx ZC706) & $\approx$ 4 \\
			ELAS-FPGA \cite{rahnama2018real} &  -- & -- & 13.6\% & -- & 0.095 & FPGA (Xilinx ZC706) & $\approx$ 3 \\
			DeepCostAggr \cite{kuzmin2017} & 5.3\% & 11.4\% & 6.3\% & 99.98\% & 0.03 & Nvidia GTX Titan X & $\approx$ 250 \\
			CSCT+SGM+MF \cite{hernandez2016embedded} & 6.9\% & 14.9\% & 8.2\% & 100\% & 0.006 & Nvidia GTX Titan X & $\approx$ 250 \\
			\bottomrule
		\end{tabular}
		\caption
		[
		Accuracy comparison of heterogenous FPGA approach to other state-of-the-art real-time methods on KITTI 2015 benchmark
		]
		{
			The quantitative results of our approach, in comparison to state-of-the-art real-time methods, on the Stereo 2015 subset of the KITTI benchmark \cite{Menze2015ISA,Menze2018JPRS}.
			As in the official evaluation protocol, we report the percentage of accurate disparities (using a threshold of $< 3$ disparity values or $5$\%, whichever is greater) after an interpolation step (meant to assign a disparity value to all pixels in the image), on respectively the subsets of background, foreground and all pixels.
			We additionally report the density of valid disparity values.
			As can be seen, with the exception of R$^3$SGM~\cite{rahnama2018fpt}, all methods provide almost dense disparity images, therefore the extra interpolation step mandated by the benchmark is not strictly required to obtain usable disparity images.
			Finally, for each method, we report the typical time required to process a stereo pair, as well as the approximate power consumption of the platform used.
			Whilst all approaches can process images in real time, only the FPGA-based methods (ours and~\cite{rahnama2018fpt}) can do so in a power-efficient manner, with ours providing $\approx 12\%$ additional accuracy and much higher density w.r.t.~\cite{rahnama2018fpt}, at the expense of slightly higher power usage and processing time.
		}
		\label{tbl:quantitativeresults-kitti-r2las}
		\vspace{-\baselineskip}
	\end{table*}

\section{Results}
\label{sec:results}

We developed the FPGA accelerators using the Vivado High-Level Synthesis (HLS) tool, as this approach was quicker, and allowed for greater flexibility and reusability of the components.
We deployed the system on a Xilinx ZCU104 board, and all of the power consumption results that we present for our method were estimated by the Xilinx Vivado tool.
Although the values provided by the tool are only approximations, they still provide an accurate sense of the power requirements.

We quantitatively evaluate the disparities produced by our approach on the standard KITTI 2015 stereo benchmark~\cite{Menze2015ISA,Menze2018JPRS}.
In Table~\ref{tbl:quantitativeresults-kitti-r2las}, we report the average percentages of pixel disparities estimated correctly for background, foreground and all pixels, respectively.
We also report average runtimes and power consumptions for both our and alternative methods that achieve real-time processing speeds on the images used in the benchmark (which have a resolution of $1242 \times 375$).
Whilst the proposed method results in slightly less accurate disparities than the DeepCostAggr~\cite{kuzmin2017} and CSCT-SGM-MF~\cite{hernandez2016embedded} methods, it is worth pointing out that both~\cite{kuzmin2017,hernandez2016embedded} rely on powerful GPUs to achieve real-time processing speed, whereas our approach does so in a much more power-efficient manner, relying only on a hybrid FPGA-CPU board.
We also compare favourably to R$^3$SGM~\cite{rahnama2018fpt}, the underlying method on which we base our approach for the estimation of the initial disparities (see \S\ref{subsec:fastr3}), providing more accurate and denser results at a similar speed and with similar power consumption.
We similarly outperform the FPGA variant of ELAS \cite{rahnama2018real}, achieving a lower error rate at a much higher speed, and with similarly low power consumption.

\begin{table}
	\centering
	  \scriptsize
	\begin{tabular}{lccc}
		\toprule
		& \textbf{Ours} & \cite{rahnama2018fpt} & \cite{rahnama2018real} \\
		\midrule
		\textbf{Platform} & ZCU104 & ZC706 & ZC706 \\
		\midrule
		\textbf{LUT Utilisation (\%~/~\#)}   & 87.5~/~$\sim$202K & 75.7~/~$\sim$165K & 37.3~/~$\sim$82K\\
		\textbf{FF Utilisation (\%~/~\#)}    & 24.1~/~$\sim$111K & 40.5~/~$\sim$177K & 22.6~/~$\sim$99K\\
		\textbf{BRAM Utilisation (\%~/~\#)}  & 70.7~/~$\sim$440  & 30.4~/~$\sim$330  & 11.9~/~$\sim$130\\
		\textbf{FPGA Power Usage (W)}  & $\approx$ 2     & 3.94         & 1.21\\
		\textbf{CPU Power Usage (W)}   & $\approx$ 3     & --           & 1.7 \\
		\textbf{Total Power Usage (W)} & $\approx$ 5     & --           & 2.91\\
		\textbf{CPU Frequency (MHz)}   & 1200 (4 Cores) & --     & 800 (2 Cores)\\
		\textbf{FPGA Frequency (MHz)}   & 200     & 100          & 100\\
		\bottomrule
	\end{tabular}
	\caption
	[
	Comparison of resource utilization and power consumption of heterogenous based approach
	]
	{
		Resources (programmable logic units), power (as estimated by the Xilinx Vivado tool) and clock frequencies used by the proposed approach, when deployed on a Xilinx ZCU104, in comparison to the FPGA-based methods from which we draw inspiration.
	}
	\label{tbl:quantitativeresults-fpga}
\end{table}

In Table~\ref{tbl:quantitativeresults-fpga}, we detail the hardware resources used by our approach when deployed on our Xilinx ZCU104 board. 
We break down the amount of logic resources used in the FPGA chip, as well as the power consumption of both the programmable logic and the ARM core.
We also report the amount of resource and power used by the methods from which we draw inspiration \cite{rahnama2018fpt,rahnama2018real}.
Notably, despite making full use of many of the logic resources available on the FPGA, our power consumption remains very low.
More specifically, breaking down the resource utilization of the programmable logic amongst the different accelerators, the largest share is taken by the Fast R$^3$SGM block which, alone, consumes about 65\% of the FPGA power.
The next most resource-heavy blocks are the ones which perform the median filtering of the disparities, which require approximately 30\% of the power.
The remaining blocks have much smaller resource requirements, which altogether account for the remaining 5\% of the power.

\section{Conclusions}

In this paper, we have presented a novel approach to computing depth from stereo images on a hybrid FPGA-CPU chip.
Our approach uses an adapted version of ELAS\ \cite{geiger2010efficient} to refine the initial sparse disparity map produced by a fast variant of R$^3$SGM \cite{rahnama2018fpt}, and achieves an 8.7\% error rate on the challenging KITTI 2015 dataset \cite{Menze2015ISA,Menze2018JPRS}.
By fully leveraging the capabilities of our hybrid board, we are able to produce highly accurate dense depth at over 50 fps, with a power consumption of only 5W, making our approach attractive for applications in mobile, real-time computing.

\section*{Acknowledgements}
\noindent
This work was supported by Innovate UK/CCAV project 103700 (StreetWise), EPSRC grant Seebibyte EP/M013774/1 and EPSRC/MURI grant EP/N019474/1.
We would also like to acknowledge the Royal Academy of Engineering and FiveAI.
\newpage
\bibliographystyle{IEEEtran}
\bibliography{refs}

\end{document}